\documentclass[]{fairmeta}
\usepackage{amsfonts}
\usepackage{adjustbox}
\usepackage{xcolor}
\usepackage{graphicx}
\usepackage{tcolorbox}
\usepackage{colortbl}
\usepackage{wrapfig}
\usepackage{tikz}

\usepackage{tcolorbox}
\tcbuselibrary{skins,breakable}
\tcbset{
  highlightbox/.style={
    enhanced,
    colback=gray!10,
    colframe=gray!50,
    boxrule=0.5pt,
    arc=2pt,
    left=6pt, right=6pt, top=6pt, bottom=6pt
  }
}

\definecolor{tokenpink}{RGB}{255, 192, 203}
\definecolor{tokengreen}{RGB}{173, 255, 47}

\newcommand{\pinksq}{\raisebox{-0.1ex}{\color{tokenpink}\rule{1.5ex}{1.5ex}}}
\newcommand{\greensq}{\raisebox{-0.1ex}{\color{tokengreen}\rule{1.5ex}{1.5ex}}}

\title{Think in Strokes, Not Pixels: Process-Driven Image Generation via Interleaved Reasoning}

\author[1,2,*]{Lei Zhang}
\author[1]{Junjiao Tian}
\author[1]{Zhipeng Fan}
\author[1]{Kunpeng Li}
\author[1]{Jialiang Wang}
\author[1]{Weifeng Chen}
\author[1]{Markos Georgopoulos}
\author[1]{Felix Juefei-Xu}
\author[3]{Yuxiang Bao}
\author[2,\dagger]{Julian McAuley}
\author[4,\dagger]{Manling Li}
\author[1,\dagger]{Zecheng He}

\affiliation[1]{Meta Superintelligence Labs}
\affiliation[2]{University of California, San Diego}
\affiliation[3]{Worcester Polytechnic Institute}
\affiliation[4]{Northwestern University}

\contribution[*]{Work done at Meta}
\contribution[\dagger]{Joint last author}

\abstract{Humans paint images incrementally: they plan a global layout, sketch a coarse draft, inspect, and refine details, and most importantly, each step is grounded in the evolving visual states. However, can unified multimodal models trained on text-image interleaved datasets also imagine the chain of intermediate states? In this paper, we introduce \textbf{process-driven image generation}, a multi-step paradigm that decomposes synthesis into an \textbf{interleaved reasoning} trajectory of thoughts and actions. Rather than generating images in a single step, our approach unfolds across multiple iterations, each consisting of 4 stages: textual planning, visual drafting, textual reflection, and visual refinement. The textual reasoning explicitly conditions how the visual state should evolve, while the generated visual intermediate in turn constrains and grounds the next round of textual reasoning. A core challenge of process-driven generation stems from the ambiguity of intermediate states: how can models evaluate each partially-complete image? We address this through dense, step-wise supervision that maintains two complementary constraints: for the visual intermediate states, we enforce the spatial and semantic consistency; for the textual intermediate states, we preserve the prior visual knowledge while enabling the model to identify and correct prompt-violating elements. This makes the generation process explicit, interpretable, and directly supervisable. To validate proposed method, we conduct experiments under various text-to-image generation benchmarks.}

\date{\today}

\begin{document}

\maketitle

\section{Introduction}
\label{sec:intro}

\begin{figure*}[ht]
    \centering
    \includegraphics[width=0.98\linewidth]{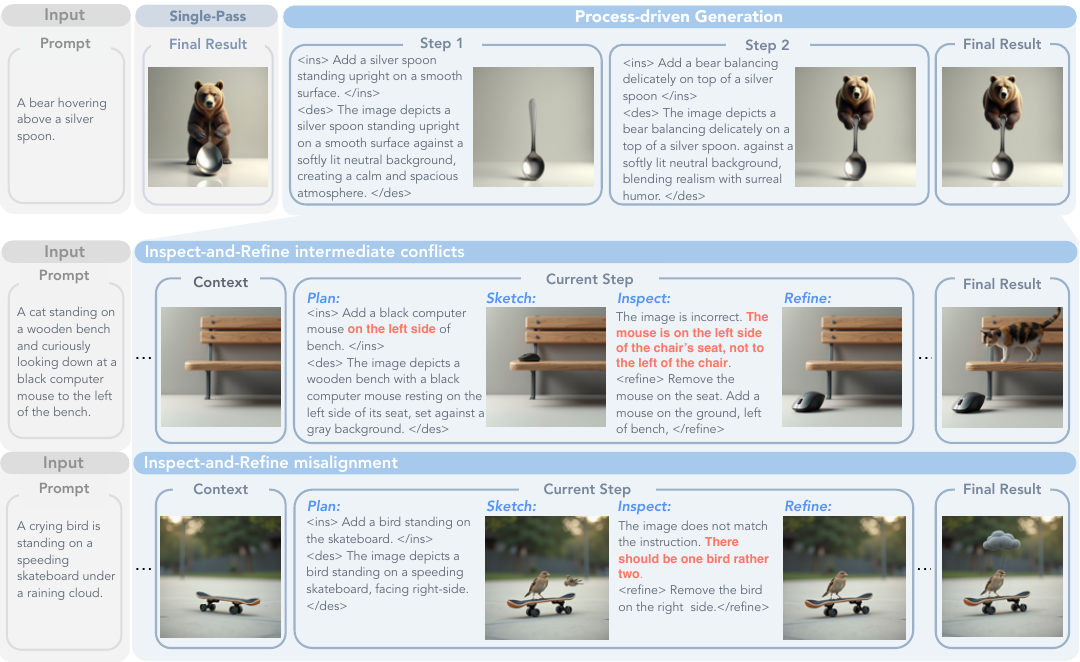}
    \caption{Single-pass generation vs process-driven interleaved reasoning. Instead of training model to generate a final image within a single pass, we teach a unified multimodal foundation model to construct the image stroke by stroke, decision by decision. This process of \texttt{Plan}, \texttt{Sketch}, \texttt{Inspect}, and \texttt{Refine} enables transformation of ambiguous intermediates into compositionally faithful final images.}
    \label{fig:overview}
\end{figure*}

Despite the impressive progress of image generation, today’s  models still remain brittle on elementary visual logic and produce plausible but incorrect images. As shown in Figure~\ref{fig:overview}, a prompt of a bear \textit{hovering above} a spoon might incorrectly yield a bear \textit{standing beside} it. Such a one-shot black-box generation forces the model to commit to an entire scene within a single forward pass, resolving precise spatial layouts, object relations, and fine-grained attributes. Although step-wise reasoning has been proposed through textual chain-of-thought (CoT)~\cite{cot,cot-2,cot-3,mvot}, but it remains visually blind, unable to dynamically perceive spatial misalignments~\cite{spatial-cot} or evolve object states~\cite{object-relation-cot,embodiedgpt}. This motivates a paradigm shift from \textit{language-driven} to genuinely \textit{multimodal, visually-grounded reasoning}. 
However, multimodal CoT~\cite{mcot,mcot-2} and tool-augmented sketching~\cite{sketchvisual,image-of-thought} decouple reasoning from generation; post-hoc refinement~\cite{unicot} remains outcome-based, where interleaving is limited to \textit{repairing after generation} rather than \textit{reasoning during generation}. 

We challenge this {outcome-driven paradigm} with \textbf{process-driven} image generation via \textbf{interleaved reasoning} anchored in both vision and text, through a unified multimodal foudnation model. We reformulate image generation with a unified multimodal models as a \textbf{co-evolving trajectory} of textual plan and visual states, orchestrated through a recurring four-stage \textbf{process}: {\texttt{Plan}} $\to$ {\texttt{Sketch}} $\to$ {\texttt{Inspect}} $\to$ {\texttt{Refine}}. The model does not hallucinate a final image; it constructs the image stroke by stroke, decision by decision. As shown in Figure~\ref{fig:overview}.,  the first stage is \texttt{Plan}, where a unified foundation model generates text instruction {\small \texttt{<ins>}} that specifies the incremental update (what to add or modify) and the state description {\small \texttt{<des>}} of the global scene hypothesis. The \texttt{Sketch} stage synthesizes a partial visual draft conditioned on prior context. The \texttt{Inspect} stage detects conflicts between sketch, plan, and prompt. The \texttt{Refine} stage produces a refinement plan {\small \texttt{<refine>}} and revise the image. This tightly-coupled loop enables {error correction as it emerges}, transforming generation from a black-box single pass into a {controllable, self-correcting dialogue} between reasoning and vision. 

A central challenge in process-driven image generation is supervising interleaved trajectories with ambiguous intermediate states, which poses three challenges:  
\textit{1) how to construct a generation path and design intermediate states}: 
Incomplete intermediate states are inherently ambiguous. For example, a missing object could be attributed to ``not yet drawn'' or ``incorrectly omitted''. We tackle this by generating multi-turn trajectories of intermediate states via \textbf{scene graph subsampling}, yielding logically ordered incremental prompts that expand the composition without contradictions.  
\textit{2) how to teach a model to {see} its mistakes}: We propose to construct a \textbf{dual-stream process-critique data} comprising error traces and alignment evaluation, to teach models to perform self-assessment from btoh textual and visual states, leveraging judges from a VLM operating on sampled trajectories.
Finally, \textit{how to jointly co-evolve text and vision reasoning}. We propose to train a \textbf{unified multimodal sequencer}, such as BAGEL~\cite{bagel}, to autoregressively generate \textit{interleaved multi-modal} tokens. As the result, the single model performs tasks decomposition, visual generation, self-validation and refinement in a sequential fashion, without the help from any external models.


Experiments prove our process-driven generation lifts the base BAGEL-7B~\cite{bagel} from $79\%$ to $83\%$ ($+4\%$) on GenEval \cite{geneval} for composition object alignment and boosts the WISE \cite{wise} for world knowledge reasoning from $70\%$ to $76\%$ ($+6\%$). Compared to existing process-level approach PARM \cite{image-cot}, our framework achieves a superior balance of performance and efficiency: it delivers higher accuracy ($0.83$ vs. $0.77$ on GenEval) with an 8x reduction in both training data and inference cost. This advantage stems from our use of semantic partitioning—supervising concrete visual states rather than the blurry latent noise used in prior work. Finally, ablation studies identify {diverse editing instructions} and {self-sampled critiques} as the key drivers of our success, outperforming symbolic corrections by $+6\%$ and proving that internalizing the model’s own failure modes is far more effective than imposing external fixes.

\section{Related Work}
\label{sec:rw}

\subsection{Unified Multimodal Model}

Unified multimodal models aim to unify visual understanding and generation within a single framework, building on the strong perceptual abilities of modern multimodal large language models. Early autoregressive approach, e.g., 
Chameleon \cite{chameleon}, Emu3 \cite{emu3}, and Show-o \cite{showo}, rely on discrete visual tokenizers such as VQ-VAE \cite{vq-vae} to model images as token sequences, achieving unified model but suffering from constrained fine-grained visual understanding. Another line of work couples a pretrained LLM with an external diffusion module, where the LLM provides semantic condition for image generation \cite{dreamllm, nextgpt, metaquery, metamorph}; while effective, this decoupled design prevents the model from fully leveraging its understanding capability during generation. More recent integrated transformer frameworks, such as the Janus series \cite{janusflow, janus-pro}, LlamaFusion \cite{llamafusion}, and BAGEL \cite{bagel}, directly combine autoregressive text modeling with diffusion-based generation to better align representation spaces and support large-scale interleaved text–image pretraining. However, despite these advances, existing unified models still struggle to tightly couple semantic reasoning with the generative process, limiting their ability to produce images with complex, logically structured content.

\subsection{Reasoning in Image Generation}

Recent studies have begun exploring interleaved reasoning in image generation, extending the success of chain-of-thought \cite{cot} from text domains \cite{cot-prompt, cot-less} to multimodal settings \cite{openaio1, vision-r1, deepseek-r1}. Early works adopt verification-based \cite{image-cot} or prompt-refinement strategies \cite{imagegen-cot}, where reasoning is performed either after image sampling or before generation; however, such text-only reasoning is isolated and remains decoupled from the evolving visual state. More recent attempts introduce multi-turn reasoning that alternates between textual analysis and visual outputs \cite{got, bagel, t2i-r1, reasongen-r1}, yet these methods typically treat images as static endpoints rather than intermediate states to be interpreted, critiqued, and updated. As a result, the reasoning flow becomes fragmented and fails to maintain coherence across steps, limiting fine-grained control over spatial relations, object dynamics, and global scene evolution. Overall, current approaches do not realize fully interleaved reasoning — where textual reasoning and visual generation mutually inform each other throughout the process—highlighting a key gap our work aims to mitigate.

\section{Method}
\label{sec:method}

\begin{figure*}[ht]
    \centering
    \includegraphics[width=\linewidth]{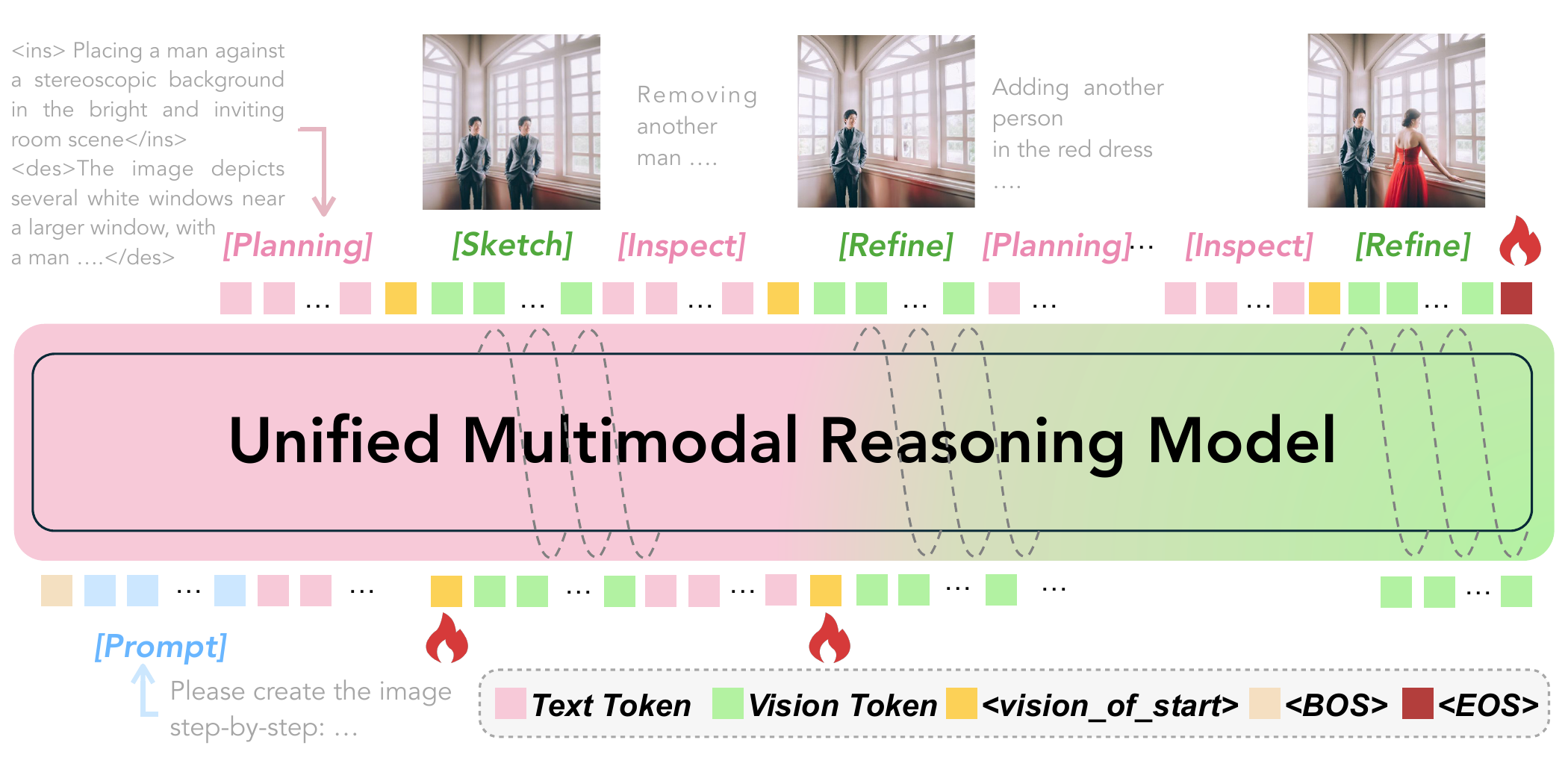}
    \caption{We design unified multimodal reasoning models for process-driven generation, autoregressively generates an interleaved sequence of \pinksq{} \textbf{text tokens} and \greensq{} \textbf{vision tokens}.}
    \label{fig:method-pipeline}
\end{figure*}

Most existing image generation models generate images in a single forward pass, sometimes augmented with chain-of-thought reasoning applied exclusively to the textual prompt. However, complex spatial relationship and fine-grained visual details are inherently difficult to encode through this one-shot paradigm, as the model must resolve the entire scene before any visual feedback is available. To overcome these limitations, we introduce process-driven image generation, which reinterprets image generation as a trajectory of interleaved intermediate states. Our method alternates between textual and visual rationales over multiple steps, enabling the model to \textbf{plan}, \textbf{sketch}, \textbf{inspect}, and \textbf{refine} the scene progressively and iteratively as generation unfolds.

\subsection{Framework}
The general framework of our model is illustrated in Figure \ref{fig:method-pipeline}. At a high level, our model performs image generation as a sequential, interleaved textual–visual reasoning process. Given a unified multimodal model $\mathcal{P}_{\theta}$ and an input text prompt $T$ (with optional input image $I_{input}$ for editing use case), the model generates a trajectory composed of alternating textual reasoning steps $s^{(i)}$ and intermediate visual states $v^{(i)}$, ultimately converging to the final image $I$:
\begin{equation}
\label{eq:method-1}
    \{ s^{(1)},v^{(1)}, s^{(2)}, v^{(2)}, ..., v^{(k)},s^{(k)} \}, I \sim \mathcal{P}_{\theta}( \cdot | T)
\end{equation}

Building on this high-level structure, the model constructs the image through a recurring four-stage cycle: Plan, Sketch, Inspect, and Refine. Each cycle incrementally advances the generation trajectory, enabling fine-grained control over both textual and visual evolution. Concretely:
\begin{itemize}
    \item \textbf{Stage I (Plan)}: The model interprets the prompt and accumulated context to produce an incremental instruction and a global scene description of the intended generation. 
    \item \textbf{Stage II (Sketch)}: Conditioned on the planned instruction, the model synthesizes an updated draft image that reflects the intended modification.
    \item \textbf{Stage III (Inspect)}: The model inspects a) the textual incremental instruction and global scene description against the raw complete prompt and b) the produced draft against the planned textual instruction to identify potential inconsistencies or mismatches.
    \item \textbf{Stage IV (Refine)}: If discrepancies are found, the model revises the instruction and generates a corrected visual update, ensuring the evolving scene remains coherent and aligned with the prompt.
\end{itemize}

Through the repeated application of these four stages, the overall generation process is decomposed into a sequence of controllable, localized updates, allowing the model to progressively assemble the final image.

Each textual intermediate $s^{(i)}$ appears in two forms depending on the stage of the cycle. During the planning phase, $s^{(i)}$ includes a step-specific painting instruction enclosed in \texttt{<ins>}...\texttt{</ins>} and a global description enclosed in \texttt{<des>}...\texttt{</des>}. During the inspection phase, if misalignment is detected, the model emits a refinement signal enclosed in \texttt{<refine>}...\texttt{</refine>}.

The corresponding visual states $v^{(i)}$ also take two forms: the planning stage produces a rough sketch that represents the intended update, while the refinement stage polishes this sketch into a more accurate visual representation. All visual outputs are wrapped between \texttt{<|vision\_start|>} and \texttt{<|vision\_end|>} to explicitly mark modality transitions.

Overall, the interleaved reasoning pipeline can be summarized as:
\begin{equation}
\label{eq:pipeline}
    T \rightarrow s^{(1)}_\text{plan} \rightarrow v^{(1)}_\text{sketch} \rightarrow s^{(1)}_\text{inspect} \rightarrow v^{(1)}_\text{refine} \rightarrow ... \rightarrow I
\end{equation}

\begin{figure*}[t]
    \centering
    \includegraphics[width=0.98\linewidth]{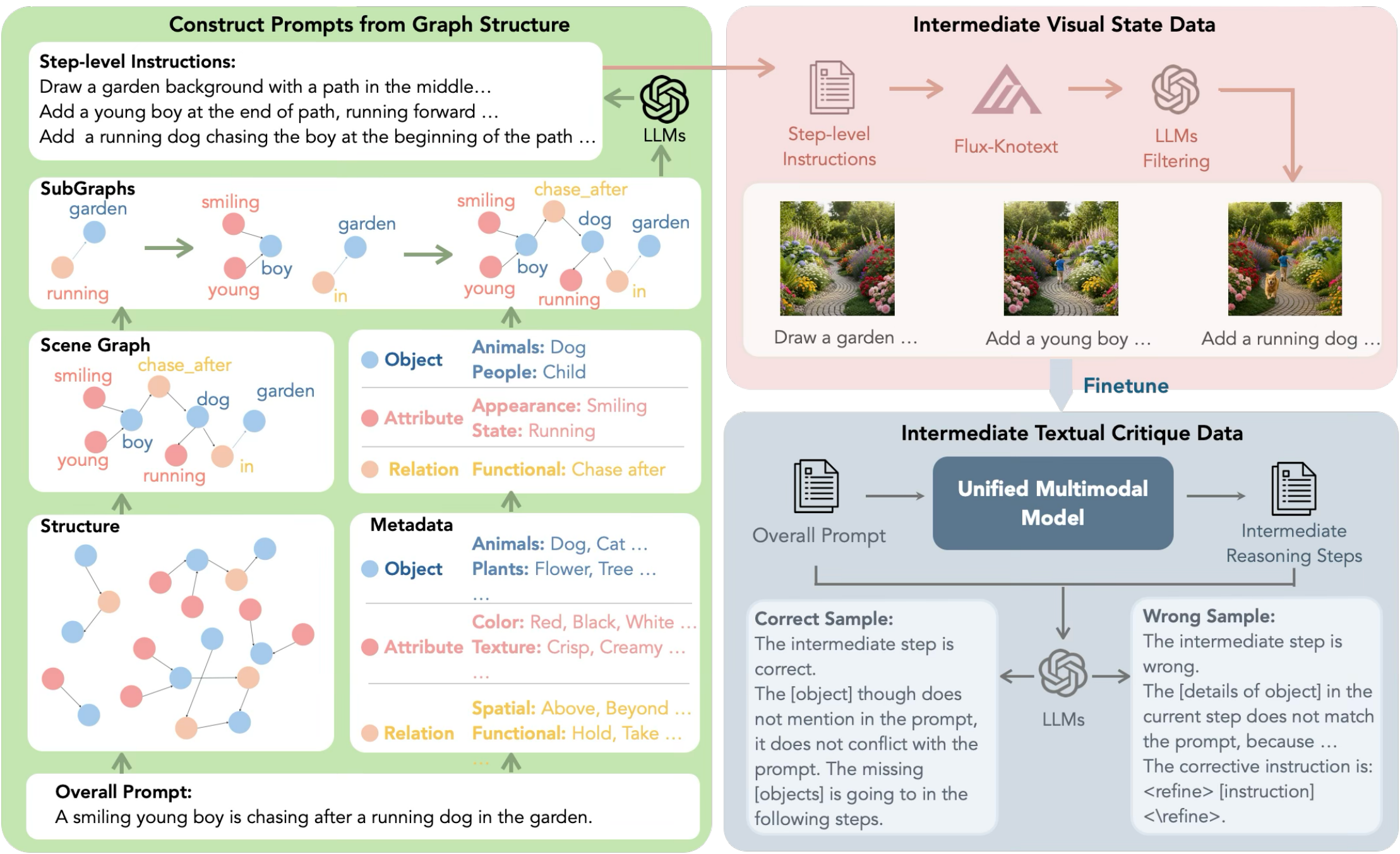}
    \caption{Our multi-stage dataset generation pipeline constructs process interleaved trajectories with intermediate visual states and textual critiques. We ensure consistent intermediate visual state generation using scene-graph structures, and generate intermediate textual critiques via self-sampling.}
    \label{fig:data-pipeline}
\end{figure*}

\subsection{Intermediate Reasoning Collection}

Advancing interleaved reasoning requires addressing a core challenge: representing and evaluating partially formed images. Unlike fully rendered outputs, intermediate visual states are inherently incomplete, making it difficult to judge whether the evolving scene preserves correct spatial structure and semantic coherence.

To equip the model with the ability to reason over such intermediate states, we introduce a multi-stage data construction pipeline that generates high-quality, process-oriented interleaved reasoning traces, as illustrated in Figure \ref{fig:data-pipeline}. We then performed supervised finetuning on this dataset to teach the model how to plan, assess, and refine visual states throughout the generation trajectory. Specifically, we introduced three dataset pipelines to comprehensively improve the models capability on planning, step-by-step generations and reasoning. We introduce them in detail below:

\paragraph{Multi-Turn Generation Subset} This dataset serves as the basic source to adapt a single-round generation model to multi-stage process-driven generation. We proposed a scene-graph based sampling mechanism to ensure at each step, only a specific local region was updated while the spatial and semantic coherence of the rest regions are preserved. Specifically, we represent each prompt using a scene graph, where object nodes, attribute nodes, and relation edges define the target composition. By subsampling subgraphs from the full scene graph, we derive a sequence of incremental step-level prompts that naturally expand the scene in a correct and controllable order — ensuring that intermediate steps remain logically grounded. We then synthesize the ground truth of each prompt using Flux-Kontext and filter with GPT. 

However, subgraph expansion alone yields primarily additive updates, which limits the diversity of the prompts. To enrich the action space of visual evolution, e.g., attribute modification, swapping, removal, we further augment it by rewriting a subset of step instructions with GPT, generating semantically equivalent but structurally different multi-step reasoning variants. This increases the coverage of visual editing behaviors and encourages the model to internalize richer transformation primitives beyond simple addition.

Finetuning with the Multi-Turn Generation subset provides the model with capabilities to perform step by step generation (i.e., performing the $s^{k}_{plan}$ and the $v^{k}_{sketch}$), but this supervision alone is insufficient for the model to generalize to high-quality reasoning. We further introduced two reasoning subsets to boost the reasoning capabilities of the model for performing $s^{k}_{inspect}$ and $s^{k}_{refine}$, which teaches the model how to interpret and reason about intermediate visual states, enabling it to distinguish valid prior visual knowledge from actual conflicts with the overall prompt. 

\paragraph{Instruction-Intermediate Conflict Reasoning Subset.} This subset focuses on improving the reasoning capabilities on the textual side. To this purpose, we adopt a self-sampling strategy: from a fine-tuned model on Multi-Turn generation Subset, we sample generated intermediate reasoning traces that include textual descriptions of partially completed images and leverage GPT as judge to eval its consistency with the original raw prompt. For conflicts, we generate a textual analysis and a corrective instruction along with the reasoning. This procedure provides explicit supervision for distinguishing incomplete-but-correct intermediate textual tokens from prompt-inconsistency reasoning.

\paragraph{Image–Instruction Alignment Reasoning Subset.} The second subset focuses on evaluating the misalignment from the visual side using the current image draft and the step-level painting instruction. We extend and refine the Gen-Ref dataset \cite{from-reflection-to-perfection} into two annotated categories: positive samples, where the image is consistent with the instruction, for which GPT generates an explanation of why the alignment holds, and negative samples, where the image mis-aligns with the instruction, for which GPT provides both an error analysis and a refinement instruction.

\begin{wraptable}{r}{0.48\linewidth}
    \vspace{-20pt}
    \centering
    \begin{adjustbox}{width=\linewidth}
    \setlength\tabcolsep{18pt}
    \setlength\extrarowheight{0.5pt}
    \arrayrulecolor[gray]{0.7}
    \begin{tabular}{lr}
    \toprule
    Statistic & Number \\
    \hline
    \rowcolor{gray!15} \multicolumn{2}{c}{\textit{Multi-turn Generation}} \\
    \hline
    Total Samples & 32,012 \\
    Average Prompt Length & 152.8 \\
    Average Image per Sample & 3.51 \\
    Maximum Image per Sample & 5 \\
    \hline
    \rowcolor{gray!15} \multicolumn{2}{c}{\textit{Instruction-Intermediate Conflict}} \\
    \hline
    Total Samples & 15,201 \\
    - Positive Samples & 6,905 \\
    - Negative Samples & 8,296 \\
    \hline
    \rowcolor{gray!15} \multicolumn{2}{c}{\textit{Image-Instruction Alignment}} \\
    \hline
    Total Samples & 15,000 \\
    - Positive Samples & 5,000 \\
    - Negative Samples & 10,000 \\
    \bottomrule
    \end{tabular}
    \end{adjustbox}
    \vspace{-5pt}
    \caption{Statistics of intermediate reasoning dataset.}
    \vspace{-40pt}
    \label{tab:dataset-stat}
\end{wraptable}

Table \ref{tab:dataset-stat} summarizes the key statistics of our curated dataset, which contains three major components tailored for process-driven image generation analysis. The Multi-turn Generation subset includes over 32K samples, each with 3–5 intermediate visual states on average, reflecting diverse multi-step reasoning trajectories. The Instruction-Intermediate Conflict subset provides more than 15K samples, covering both positive and negative cases to supervise the model’s ability to detect and correct instruction conflicts. Additionally, the Image-Instruction Alignment subset comprises 15K samples with balanced positive and negative examples to support evaluation of fine-grained visual-text alignment.

\subsection{Model}

To enable process-driven interleaved reasoning, the model must possess unified multimodal understanding and capabilities. Therefore, we adopt a unified multimodal model, such as BAGEL \cite{bagel}, as our backbone and finetune it for our interleaved process driven generation task. 

\paragraph{Training Objectives.} We train our model to generate text tokens autoregressively, optimizing the process with a Cross-Entropy (CE) Loss. The loss is applied only to positions corresponding to textual segments $s^{(i)}$.
In order to natively generate interleaved sequences, we add a loss term on the \texttt{<|vision\_start|>} and \texttt{<|vision\_end|>} tokens, enabling the seamless switch between textual and visual tokens generation. The CE loss for next-token prediction is formally defined as,
\begin{equation}
\label{eq:ce-loss}
    \mathcal{L}^{\text{text}}_{\mathrm{CE}} = - \sum_{t \in [1,i]} \log \mathcal{P}_{\theta} \big( s_{t} \;\, | \;\, y_{<t}, T \big)
\end{equation}
On the visual side, we followed \cite{bagel} and employ Rectified Flow paradigm \cite{flow} to generate images, following:
\begin{equation}
\label{eq:latent-eq}
    z_{t}^{(i)} = t \cdot z_{0}^{(i)} + (1-t)\cdot z_{1}^{(i)}, \quad t \in [0,1]
\end{equation}
\begin{equation}
\label{eq:mse-loss}
\mathcal{L}^{\text{image}}_{\mathrm{MSE}} = \mathbb{E} \!\left[
    \left\lVert
      \mathcal{P}_{\theta}\!\big( z_t^{(i)} \mid y_{<t}, T \big)
      - (z_{0}^{(i)} - z_{1}^{(i)})
    \right\rVert^{2}
  \right],
\end{equation}
The model conditions on entire preceding contexts, consisting of the input prompt and the chain of interleaved reasoning trajectories (including visual components, corresponding textual guidance, etc). The overall object for our training is a weighted combination of the above objectives.
\begin{equation}
\label{eq:total-loss}
\mathcal{L}_{\text{total}} = \lambda_{\mathrm{CE}} \cdot \mathcal{L}^{\text{text}}_{\mathrm{CE}}
  + \mathcal{L}^{\text{image}}_{\mathrm{MSE}}
\end{equation}
where the hyperparameter $\lambda_{CE}$ is a scaling coefficient to balance CE loss and MSE loss. 

\paragraph{Inference.} During inference, given a textual prompt $T$, the model autoregressively generates an interleaved reasoning trajectory. Textual and visual intermediates are produced in a unified sequence, where modality shifts are governed by special tokens. The whole process terminates once the the final completed image $I$ is emitted, marked by an end-of-sequence token.
\section{Experiment}
\label{sec:exp}

\begin{table*}[]
    \centering
    \setlength\tabcolsep{9pt} 
    \setlength\extrarowheight{0pt} 
    \arrayrulecolor[gray]{0.7} 
    \begin{adjustbox}{width=\linewidth}
    \begin{tabular}{ccccccc|c}
    \toprule
    {\textbf{Model}} & {\textbf{Single Object}} & {\textbf{Two Objects}} & {\textbf{Counting}} & {\textbf{Colors}} & {\textbf{Position}} & {\textbf{Color Attributes}} & \textbf{Overall}$\uparrow$ \\ 
    \hline
    \rowcolor{gray!15} \multicolumn{8}{c}{\textit{Generation Only}} \\
    \hline
    PixArt-$\Sigma$ \cite{pixartsigma} & 0.98 & 0.50 & 0.44 & 0.80 & 0.08 & 0.07 & 0.48 \\
    SDv2.1 \cite{sd21} & 0.98 & 0.51 & 0.44 & 0.85 & 0.07 & 0.17 & 0.50 \\
    DALL-E 2 \cite{dalle2} & 0.94 & 0.66 & 0.49 & 0.77 & 0.10 & 0.19 & 0.52 \\
    Emu3-Gen \cite{emu3} & 0.98 & 0.71& 0.34 & 0.81 & 0.17 & 0.21 & 0.54 \\
    SDXL \cite{sdxl} & 0.98 & 0.74 & 0.39 & 0.85 & 0.15 & 0.23 & 0.55 \\
    DALL-E 3 \cite{dalle3} & 0.96 & 0.87 & 0.47 & 0.83 & 0.43 & 0.45 & 0.67 \\
    SD3\hbox{-}Medium \cite{sd3medium} & 0.99 & 0.94 & 0.72 & 0.89 & 0.33 & 0.60 & 0.74 \\
    FLUX.1-dev (12B) \cite{flux-kontext} & 0.98 & 0.93 & 0.75 & 0.93 & 0.68 & 0.65 & 0.82 \\
    \hline
    \rowcolor{gray!15} \multicolumn{8}{c}{\textit{Unified Multimodal}} \\
    \hline
    Chameleon \cite{chameleon} & {--} & {--} & {--} & {--} & {--} & {--} & 0.39 \\
    LWM \cite{lwm} & 0.93 & 0.41 & 0.46 & 0.79 & 0.09 & 0.15 & 0.47 \\
    SEED\hbox{-}X \cite{seedx} & 0.97 & 0.58 & 0.26 & 0.80 & 0.19 & 0.14 & 0.49 \\
    TokenFlow\hbox{-}XL \cite{tokenflowxl} & 0.95 & 0.60 & 0.41 & 0.81 & 0.16 & 0.24 & 0.55 \\
    ILLUME \cite{Illume} & 0.99 & 0.86 & 0.45 & 0.71 & 0.39 & 0.28 & 0.61 \\
    Transfusion \cite{transfusion} & {--} & {--} & {--} & {--} & {--} & {--} & 0.63 \\
    Emu3\hbox{-}Gen \cite{emu3} & 0.99 & 0.81 & 0.42 & 0.80 & 0.49 & 0.45 & 0.66 \\
    Janus \cite{janus} & 0.97 & 0.68 & 0.30 & 0.84 & 0.46 & 0.42 & 0.61 \\
    Janus\hbox{-}Pro\hbox{-}7B \cite{janus-pro} & 0.99 & 0.89 & 0.59 & 0.90 & 0.79 & 0.66 & 0.80 \\
    Show\hbox{-}o \cite{showo} & 0.98 & 0.80 & 0.66 & 0.84 & 0.31 & 0.50 & 0.68 \\
    Show\hbox{-}o2 \cite{showo2} & 1.00 & 0.87 & 0.58 & 0.92 & 0.52 & 0.62 & 0.76 \\
    BAGEL-7B$^{*}$ \cite{bagel} & 0.99 & 0.95 & 0.76 & 0.87 & 0.51 & 0.56 & 0.77 \\
    \midrule
    Ours (BAGEL-7B + Process-Driven) & 0.99 & 0.95 & 0.75 & 0.87 & 0.72 & 0.69 & \textbf{0.83}\\
    \bottomrule
    \end{tabular}
     \end{adjustbox}
    \caption{\textbf{Evaluation of text-to-image generation ability on Gen-Eval benchmark.} ``Generation Only'' stands for an image generation model, and ``Unified Multimodal'' denotes a model that has both understanding and generation capabilities. $^{*}$ means we report the
reproducing results using the official Github repository and checkpoint. Our approach boosts BAGEL-7B by 5\% absolute gains, and the performance with 7B parameters achieve comparable with a 12B state-of-the-art generation-only model FLUX.1-dev. }
    \label{tab:quantitative-result}
\end{table*}

\begin{table*}[ht]
    \centering
    \setlength\tabcolsep{12pt} 
    \setlength\extrarowheight{0pt} 
    \arrayrulecolor[gray]{0.7} 
     \begin{adjustbox}{width=\linewidth}
    \begin{tabular}{cccccccc}
    \toprule
    \textbf{Model} & \textbf{Culture} & \textbf{Time} & \textbf{Space} & \textbf{Biology} & \textbf{Physics} & \textbf{Chemistry} & \textbf{Overall}$\uparrow$ \\
    \hline
    \rowcolor{gray!15} \multicolumn{8}{c}{\textit{Generation Only}} \\
    \hline
    SDv1.5 \cite{sd1.5} & 0.34 & 0.35 & 0.32 & 0.28 & 0.29 & 0.21 & 0.32 \\
    SDXL  \cite{sdxl} & 0.43 & 0.48 & 0.47 & 0.44 & 0.45 & 0.27 & 0.43 \\
    SD3.5-large \cite{sd3medium} & 0.44 & 0.50 & 0.58 & 0.44 & 0.52 & 0.31 & 0.44 \\
    PixArt-$\Sigma$ \cite{pixartsigma} & 0.45 & 0.50 & 0.48 & 0.49 & 0.56 & 0.34 & 0.47 \\
    Playground-v2.5 \cite{playground-v2.5} & 0.49 & 0.58 & 0.55 & 0.43 & 0.48 & 0.33 & 0.49 \\
    FLUX.1-dev \cite{flux-kontext} & 0.48 & 0.58 & 0.62 & 0.42 & 0.51 & 0.35 & 0.50 \\
    \hline
    \rowcolor{gray!15} \multicolumn{8}{c}{\textit{Unified Multimodal}} \\
    \hline
    Janus \cite{janus} & 0.16 & 0.26 & 0.35 & 0.28 & 0.30 & 0.14 & 0.23 \\
    VILA-U \cite{vila-u} & 0.26 & 0.33 & 0.37 & 0.35 & 0.39 & 0.23 & 0.31 \\
    Show-o \cite{showo} & 0.28 & 0.40 & 0.48 & 0.30 & 0.46 & 0.30 & 0.35 \\
    Janus-Pro-7B \cite{janus-pro} & 0.30 & 0.37 & 0.49 & 0.36 & 0.42 & 0.26 & 0.35 \\
    Emu3 \cite{emu3} & 0.34 & 0.45 & 0.48 & 0.41 & 0.45 & 0.27 & 0.39 \\
    MetaQuery \cite{metaquery} & 0.56 & 0.55 & 0.62 & 0.49 & 0.63 & 0.41 & 0.55 \\
    Show-o2 \cite{showo2} & 0.64 & 0.58 & 0.61 & 0.58 & 0.63 & 0.49 & 0.61 \\
    BAGEL \cite{bagel} & 0.76 & 0.69 & 0.75 & 0.64 & 0.75 & 0.58 & 0.70 \\
    \midrule
    Ours (BAGEL + Process-driven) & 0.74 & 0.82 & 0.73 & 0.70 & 0.76 & 0.78 & \textbf{0.76} \\
    \bottomrule
    \end{tabular}
    \end{adjustbox}
    \caption{\textbf{Evaluation of world knowledge reasoning WISE benchmark.} WISE assesses a model’s ability to integrate world knowledge and structured semantic reasoning into text-to-image generation. ``Generation Only" stands for an image genreation model, and ``Unified Multimodal" models jointly support both understanding and generation. Our approach boosts BAGEL-7B by 8.5\% absolute gains, achieving nearly 15\% gains on challenging tasks like Time and Chemistry.}
    \label{tab:wise}
\end{table*}

\subsection{Implementation Details}

\paragraph{Dataset.} We construct a process-based interleaved reasoning dataset from scratch, consisting of three complementary components: the multi-turn generation subset, instruction-intermediate conflict set and image-instruction alignment set. The multi-turn generation subset contains 30K samples, each paired with approximately three intermediate images and corresponding step-level instructions. To support prompt–intermediate conflict reasoning, we further collect 15K samples by self-sampling intermediate trajectories as the instruction-intermediate conflict set. Finally, the image–instruction alignment subset includes 15K image–text pairs (10K negative and 5K positive) to enforce fine-grained consistency between editing instructions and resulting visual modifications. Additional dataset construction details are provided in Sec.~\ref{sec:method}.

\paragraph{Training and Inference Details.} We adopt the unified multimodal understanding and generation model BAGEL-7B \cite{bagel} as our base model. Throughout the training, all the model parameters are finetuned end-to-end on a node with 8 NIVIDA H100 GPUs for 10,000 steps using a packed sequence of 33,000 tokens, a learning rate of $2 \times 10^{-5}$, and cosine decay. We extend the original training objective to support seamless transitions between textual reasoning and visual generation within a single autoregressive sequence. At inference time, when encounter \texttt{<|vision\_start|>}, the model seamlessly switches to image generation mode to generate visual aids. The entire interleaved generation process only stops if the model generates the \texttt{<vision\_end>} without \texttt{<|vision\_start|>} following.

\subsection{Quantitative Evaluation}

Table \ref{tab:quantitative-result}. demonstrates the quantitative results on the GenEval benchmark \cite{geneval}, which evaluates compositional text-to-image in various object-centric attributes. The evaluation includes both generation-only models and unified multimodal models. Our method exhibits particularly large gains on relational and attribute-sensitive tasks, such as position and color attribute. These categories require precise spatial reasoning and fine-grained cross-model alignment, which single-pass generative models frequently fail to capture and unified multimodal models often struggle to integrate coherently. In contrast, our process-driven interleaved reasoning yields more reliable object grounding and attribute consistency, enabling our model to match or surpass the best-performing unified systems (e.g., Janus-Pro-7B and BAGEL) while maintaining strong performance on simple single-object cases. Overall, our method sets a new state of the art among unified models, demonstrating that interleaving text–visual reasoning substantially enhances structural fidelity in image generation.

Table \ref{tab:wise}. demonstrates the quantitative results on the WISE \cite{wise} benchmark, which is designed to assess world knowledge reasoning in text-to-image generation. Generation-only models achieve moderate performance overall, ranging from 0.32 to 0.50, due to their limited multimodal understanding capabilities. Unified multimodal models, such as Janus-Pro and BAGEL, exhibit stronger results. However, these models struggle with temporal and scientific domains, such as Time and Chemistry. By incorporating process-driven reasoning, our model achieves the best overall score 0.76 and delivers consistent improvements across nearly all domains. In particular, we observe substantial gains in Time, Biology, and Chemistry, demonstrating enhanced generalization to complex concepts. These results show that interleaved reasoning trajectories enable models to better utilize world knowledge during generation.

\begin{table*}[t]
    \centering
    \setlength\tabcolsep{5pt} 
    \setlength\extrarowheight{0pt} 
    \arrayrulecolor[gray]{0.7} 
     \begin{adjustbox}{width=\linewidth}
    \begin{tabular}{cccc}
    \toprule
    \textbf{Reasoning Strategy} & \textbf{Training Dataset} & \textbf{Inference Cost} & \textbf{Gen-Eval Bench} \\
    \hline
    \rowcolor{gray!15} \multicolumn{4}{c}{\textit{Training-free}} \\
    \hline
    BAGEL \cite{bagel} + GPT (Planner) & - & 50 & 0.60 \\
    BAGEL \cite{bagel} + GPT (Inspector) & - & 50 & 0.80 \\
    \hline
    \rowcolor{gray!15} \multicolumn{4}{c}{\textit{Training-based}} \\
    \hline
    PARM \cite{image-cot} (TTS) & 400K & 1000 & 0.67 \\
    PARM \cite{image-cot} (RL + TTS) & 688K & 1000 & 0.77 \\
    \midrule
    Ours (SFT) & \textbf{62K} & \textbf{131} & \textbf{0.83} \\
    \bottomrule
    \end{tabular}
    \end{adjustbox}
    \caption{\textbf{Qualitative comparison with process-driven baselines on Gen-Eval benchmark.} ``Training Dataset'' indicates the number of samples used for fine-tuning. ``Inference Cost'' denotes the cumulative of sampling steps required to synthesize a final image. Our method achieves superior performance with significantly lower training and inference overhead compared to existing process-based approaches.}
    \label{tab:process-eval}
\end{table*}

\subsection{Analysis of Process-driven Reasoning}

To evaluate the effectiveness of our proposed paradigm, we conduct a comparative analysis against several process-based baselines. This experiment aims to verify whether our model effectively internalizes visual-semantic grounding and to demonstrate its efficiency compared to existing training-free and training-based workflows.

We evaluate our approach against two distinct categories of baselines: 
\begin{itemize}
    \item \textbf{BAGEL + GPT (Planner):} Use GPT-4o as external guidance to provide step-by-step instructions for multi-turn image generation by the vanilla BAGEL model.
    \item \textbf{BAGEL + GPT (Inspector):} Employ BAGEL to perform a single-round refinement of its initial draft conditioned on verbal feedback regarding inconsistencies provided by a GPT-4o.
    \item \textbf{PARM:} A state-of-the-art baseline that performs step-level verification by assessing the potential of intermediate, blurry decoding states to satisfy the prompt, used for path selection (TTS) or iterative alignment (RL).
\end{itemize}

The comparison with training-free methods in Table \ref{tab:process-eval}. underscores that \textbf{SFT is indispensable for internalizing visual-semantic grounding.} Attempting to use GPT-4o as an external planner for the base model leads to a 23\% performance collapse. This occurs because, without specialized tuning, the base model cannot consistently ground incremental, multi-step instructions into stable visual intermediates. Furthermore, external verbal critiques from an inspector provide only marginal gains, confirming that verbal feedback cannot be effectively translated into precise visual updates unless the model's native generative path is aligned through process-level training. Our approach bridges this gap, enabling a 7B model to surpass the performance of the 12B FLUX.1-dev and even surpass it by 26\% on the WISE benchmark.

Compared to training-based baselines PARM, our method demonstrates a superior balance between performance and efficiency across multiple dimensions:
\begin{itemize}
\item \textbf{Computational and Data Efficiency:} ur model establishes a new SOTA score of 0.83 on GenEval, outperforming even the RL-enhanced PARM (0.77). Crucially, we achieve this using only \textbf{62K} samples—an $11\times$ reduction in training data compared to PARM’s 688K. During inference, our model provides a nearly $8\times$ speedup. While PARM relies on a Best-of-20 search strategy that incurs a cumulative cost of 1000 sampling steps, our model synthesizes high-quality images in just 131 steps (averaging 2.62 reasoning steps per image). This inference cost is calculated based on our model's complexity-adaptive reasoning, where it autonomously determines the trajectory length based on task difficulty.
\item \textbf{Visual-Semantic Partitioning vs. Pixel-level Supervision:} A fundamental distinction lies in the nature of the "process" itself. PARM supervises the diffusion latent space, where intermediate states are often fuzzy or lack clear human-interpretable meaning. In contrast, we utilize semantic partitioning, where reasoning trajectories are adaptively driven by the emergence of objects and relations. By supervising the model on trajectories with concrete visual semantics rather than abstract noise levels, our model achieves superior grounding and complexity-adaptive reasoning.
\end{itemize}

\begin{table}[t]
    \centering
    \begin{minipage}[ht]{0.48\textwidth}
    \centering
    \begin{adjustbox}{width=\linewidth}
    \setlength\tabcolsep{6pt}
    \setlength\extrarowheight{0pt}
    \begin{tabular}{lccc}
        \toprule
        \textbf{Case} & \textbf{Color} & \textbf{Position} & \textbf{Color Attr.} \\
        \midrule
        w/o aug. & 0.81 & 0.58 & 0.50 \\
        + Self-critique & 0.84 & 0.61 & 0.53 \\
        \midrule
        w/ aug.  & 0.82 & 0.67 & 0.62 \\
        + Self-critique & \textbf{0.87} & \textbf{0.72} & \textbf{0.69} \\
        \bottomrule
    \end{tabular}
    \end{adjustbox}
    \caption{Diverse editing instructions unlock relational reasoning: augmenting additive step prompts with richer operations (\textit{refine, remove, swap}) boosts performance.}
    \label{tab:ablation-gen-data}
    \end{minipage}
    \hfill
    \begin{minipage}[ht]{0.48\textwidth}
    \centering
    \begin{adjustbox}{width=\linewidth}
    \setlength\tabcolsep{6pt}
    \setlength\extrarowheight{0pt}
    \begin{tabular}{lccc}
        \toprule
        \textbf{Case} & \textbf{Color} & \textbf{Position} & \textbf{Color Attr.} \\
        \midrule
         & 0.82 & 0.67 & 0.62 \\
        + scene graph & 0.83 & 0.70 & 0.67 \\
        + self-sampling & \textbf{0.87} & \textbf{0.72} & \textbf{0.69} \\
        \bottomrule
    \end{tabular}
    \end{adjustbox}
    \caption{Supervising refinement via the model’s own error trajectories (self-sampling) yields better performance over scene-graph-derived corrections.}
    \label{tab:ablation-critique}
    \end{minipage}
\end{table}

\begin{table}[t]
    \centering
    \centering
    \begin{adjustbox}{width=\linewidth}
    \setlength\tabcolsep{8pt}
    \setlength\extrarowheight{0pt}
    \begin{tabular}{ccccc}
    \toprule
    \textbf{Case} & \textbf{Counting} & \textbf{Colors} & \textbf{Position} & \textbf{Color Attr.} \\
    \midrule
         &  0.61     & 0.84   & 0.66     & 0.62 \\
    + Instruction-intermediate conflict (w/ ins.) & 0.62 & 0.85 & 0.71 & 0.65 \\
    + mage-Instruction alignment (w/ img-ins.) & 0.73 & 0.86 & 0.69 & 0.65 \\
    w/ ins. + img-ins. (ours) & \textbf{0.75} & \textbf{0.87} & \textbf{0.72} & \textbf{0.69} \\
    \bottomrule
    \end{tabular}
    \end{adjustbox}
    \caption{\textbf{Complementary designs on intermediates states improve different generation tasks:} Instruction-intermediate conflict supervision (w/ ins.) primarily improves semantic and spatial consistency (e.g., Position), while Image-Instruction alignment supervision (w/ img-ins.) yields gains in visually grounding (e.g., Counting).}
    \label{tab:intermediate-abs}
\end{table}

\subsection{Ablation Study and Analysis}

To understand the contribution of each component in our process-driven recipe, we conduct an extensive ablation study on the GenEval benchmark across three dimensions: the structure of step-level instructions, the construction of intermediate supervision, and the design of semantic–visual consistency checks. The results show that intermediate reasoning quality emerges from the interaction of these components rather than any single factor.

\textbf{Diverse step-level instructions enable model to develop more flexible intermediate reasoning strategies.} To investigate how the form of step-level instructions influences multi-turn generation, we compare two variants of our multi-turn generation dataset. The first variant constructs instructions from incrementally expanded subgraphs of the scene graph, resulting in a sequence of purely additive editing operations. The second variant augments this dataset by rewriting a subset of instructions using GPT, introducing a richer set of editing types such as attribution modification, swapping, removal, etc.

As shown in Table~\ref{tab:ablation-gen-data}., the dataset with only additive operations (w/o aug.) achieves moderate performance consistently. Introducing instruction diversity (w/ aug.) leads to clear improvements, especially in position (+0.09) and attribute accuracy (+0.12). After applying self-critique fine-tuning, both settings improve, but the diversified-instruction variant again shows a larger boost (+0.11 in position, +0.16 in attributes), demonstrating that models benefit from learning richer forms of intermediate visual reasoning.

This suggests that intermediate process supervision benefits from exposing the model to varied forms of visual reasoning, enabling it to better interpret evolving visual states and execute more flexible and coherent edits, rather than following a single monotonic editing trajectory. We attribute the improvement to the role of instruction diversity in shaping the model’s intermediate reasoning behavior, which mirrors real-world creative workflows. Such diversity encourages the model to interpret partially formed images not simply as ``incomplete additions", but as editable states that may require correction or adjustment, which also works as the basis of further inspecting and refining. 

\textbf{The consistency of critique space is more important than the controability.} To investigate the optimal construction of intermediate critique supervision, we compare two data generation strategies: Symbolic Corrections, which derives corrections from the scene graph: we target a specific object, attribute, or relation and produce a corresponding critique, and Self-Sampling, where GPT identifies alignment conflicts within model-generated visual trajectories and proposes corrective instructions.

Table \ref{tab:ablation-critique}. compares different strategies for constructing intermediate critique supervision. Starting from the baseline multi-turn model (Color: 0.82, Position: 0.67, Attribute: 0.62), incorporating scene graph–based critiques leads to moderate improvements (0.83 / 0.70 / 0.67), suggesting that explicitly correcting individual objects, attributes, or relations can help guide the refinement process. However, replacing these symbolic edits with self-sampled critiques yields substantially larger gains (0.87 / 0.72 / 0.69).

We attribute this improvement to the fact that self-sampling operates in the model’s own distribution. The critique data reflects the model’s actual failure modes and correction needs, enabling the supervision signal to be aligned with the model’s internal reasoning dynamics.

\begin{figure*}[t]
    \centering
    \includegraphics[width=0.98\linewidth]{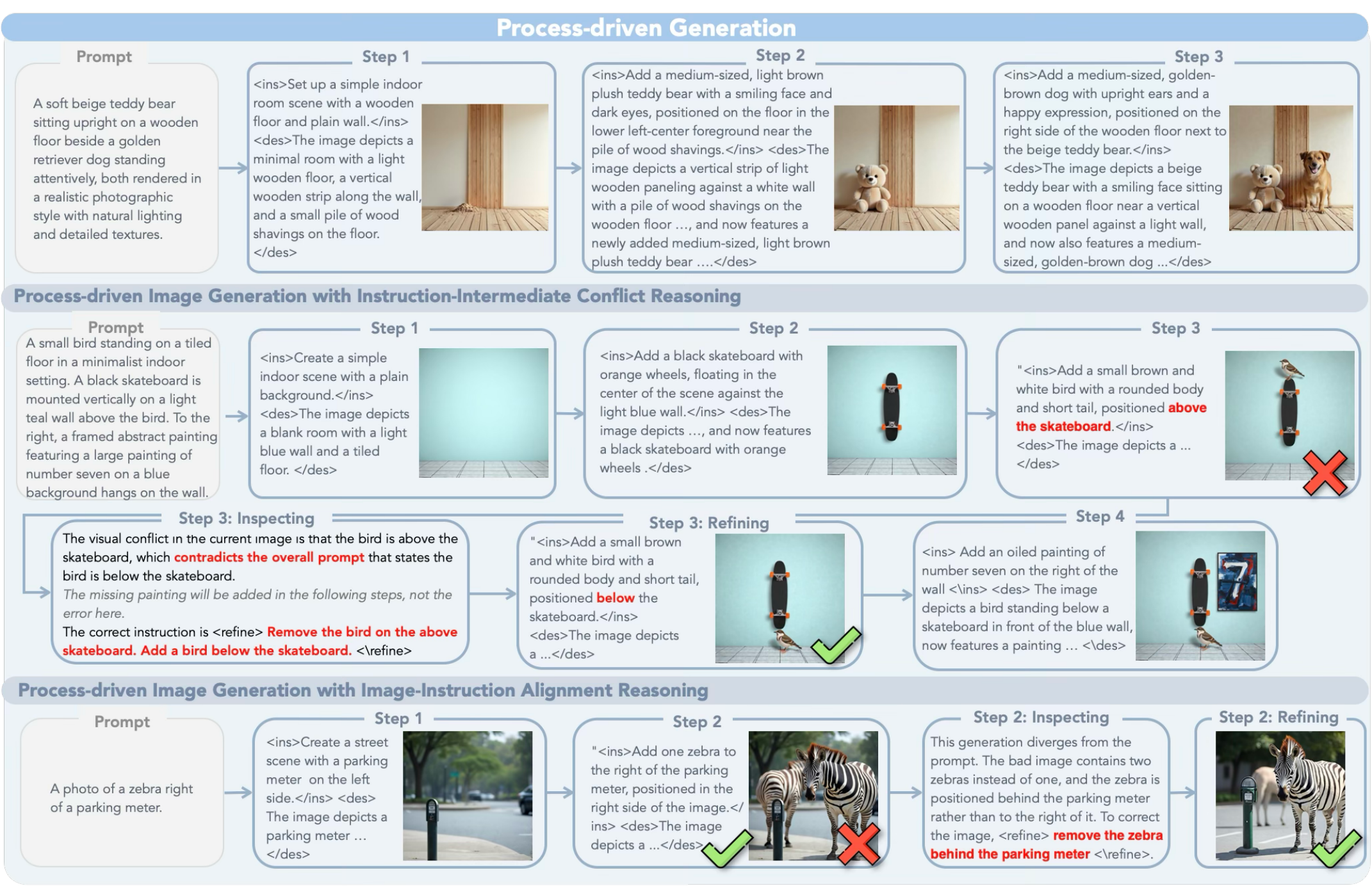}
    \caption{\textbf{Visualization of the interleaved reasoning trajectory in our process-driven image generation.} Each step follows the plan–sketch–inspect–refine cycle, while \texttt{inspect} steps with no detected issues are omitted for brevity. The second and third rows illustrate two types of intermediate errors: (1) conflicts between the step-level instruction and the overall prompt, where the model revises the instruction and corrects the image; and (2) inconsistencies between the generated draft and instruction, where the instruction and overall prompt remain valid but the image requires refinement.}
    \label{fig:visualization}
\end{figure*}

\textbf{Semantic- and visual-level constraints address distinct failure modes and jointly improve compositional reasoning accuracy.} Table \ref{tab:intermediate-abs} analyzes two complementary mechanisms that intervene on intermediate reasoning states. Adding instruction-intermediate conflict supervision (w/ ins.) improves semantic and spatial consistency, evidenced by the gains on the Position category (+5\%). This mechanism helps ensure that the evolving plan remains semantically aligned with the global intent, thereby reducing instruction drift and preserving correctness throughout the reasoning trajectory, which directly translates to better grounding in spatial and attribute-centric tasks. On the other hand, image-instruction alignment supervision (w/ img-ins.) sharpens visually grounded reasoning, producing substantial gains on Counting (+12\%). Notably, combining both mechanisms achieves the highest performance across all tasks, revealing that semantic-level and visual-level intermediate checks target distinct and complementary failure modes. These results establish that enforcing correctness during intermediate steps—not merely at the final output—is essential for reliable multi-step generation, enabling the model to actively detect and rectify inconsistencies before they propagate.

\subsection{Qualitative Evaluation}

\begin{figure*}[!t]
    \centering
    \includegraphics[width=0.98\textwidth]{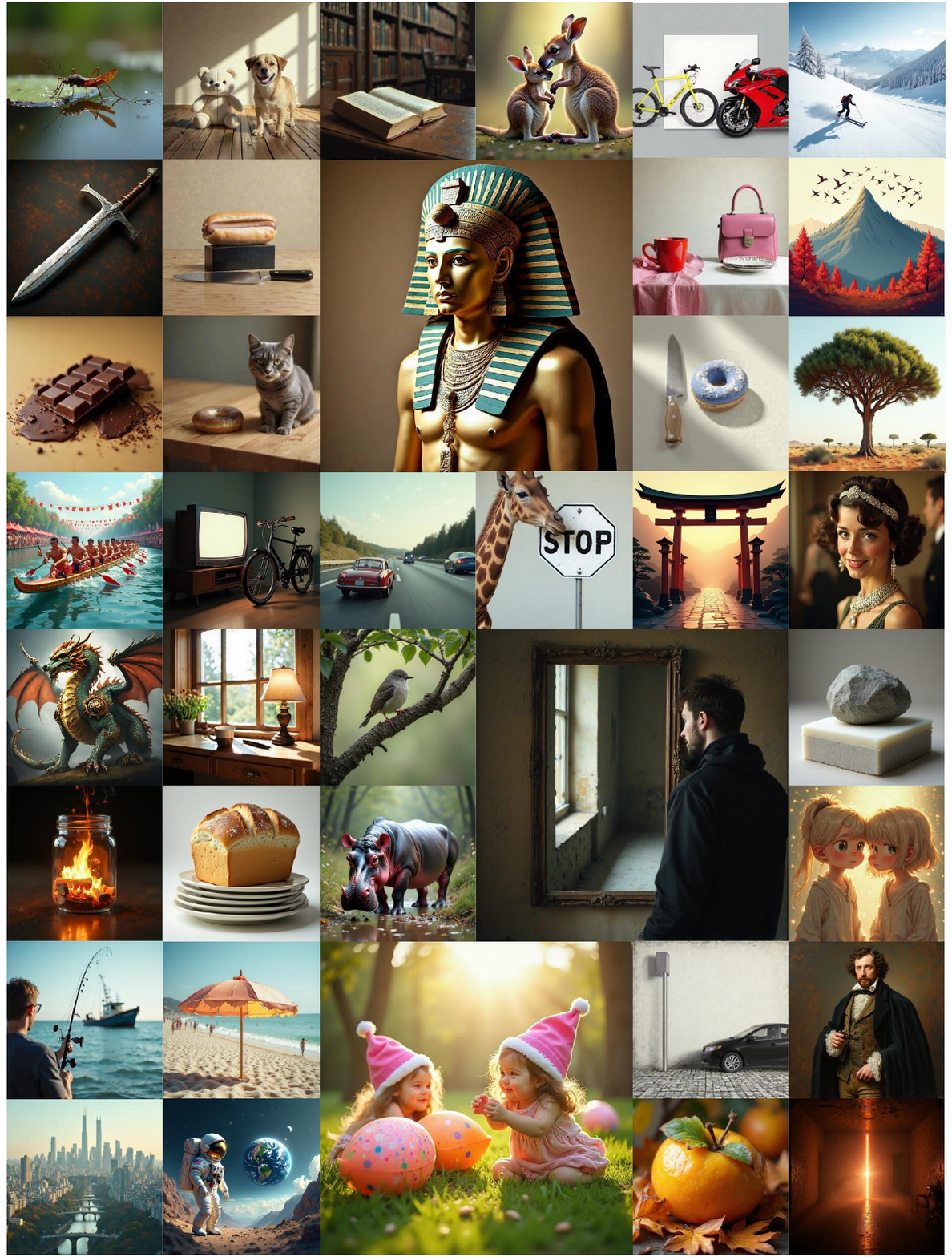}
    \caption{\textbf{Visualization of generated image in our process-driven image generation.} Our process-driven approach produces images with high visual fidelity, fine-grained details, and strong aesthetic appeal. The prompts are sampled from Gen-Eval and WISE benchmark.}
    \label{fig:vis-supp}
\end{figure*}

Figure \ref{fig:visualization}. illustrates the reasoning trajectories produced by our process-driven generation paradigm. The first row shows that the model transforms conventional single-pass generation into a sequence of adaptive reasoning steps, progressively refining both the textual plan and the visual draft. The second and third rows highlight the model’s ability to detect and correct two distinct types of intermediate errors: (1) conflicts between the step-level instruction and the overall prompt, and (2) mismatches between the instruction and the partially generated image. These complementary mechanisms enable the model to revise wrong intermediate states and supply a coherent context for subsequent updates. Notably, the model does not misinterpret incomplete or yet-to-be-rendered details as errors, demonstrating its capacity to distinguish intermediate progress from true inconsistencies. As shown in Figure \ref{fig:vis-supp}., our process-driven approach produces images with high visual fidelity, fine-grained details, and strong aesthetic appeal. The prompts are sampled from Gen-Eval and WISE benchmark.

\section{Conclusion}
\label{sec:conclusion}


We introduce a novel process-driven interleaved reasoning paradigm that teaches a unified multimodal model to build images stroke by stroke, decision by decision, via a co-evolving loop of textual planning, visual sketching, self-inspection, and refinement. Our method hinges on three breakthroughs: scene-graph subsampling for contradiction-free incremental instructions, self-sampled critique traces to learn from the model’s errors, and end-to-end training of BAGEL-7B to autoregressively emit interleaved text and image tokens.  
We lift the public BAGEL-7B from 0.79 to 0.83 (+5\% absolute gain) on GenEval and from 0.70 to 0.76 (+6\% absoluate gain) on WISE. 
Looking forward, we will extend unified multimodal reasoning to videos and 3D space, and enable real-time human-in-the-loop control, unlocks controllable, truthful and interpretable image synthesis.

\clearpage
\newpage
\bibliographystyle{assets/plainnat}
\bibliography{paper}

\begin{thebibliography}{55}
\providecommand{\natexlab}[1]{#1}
\providecommand{\url}[1]{\texttt{#1}}
\expandafter\ifx\csname urlstyle\endcsname\relax
  \providecommand{\doi}[1]{doi: #1}\else
  \providecommand{\doi}{doi: \begingroup \urlstyle{rm}\Url}\fi

\bibitem[Betker et~al.()Betker, Goh, Jing, TimBrooks, Wang, Li, LongOuyang, JuntangZhuang, JoyceLee, YufeiGuo, WesamManassra, PrafullaDhariwal, CaseyChu, YunxinJiao, and Ramesh]{dalle3}
James Betker, Gabriel Goh, Li~Jing, † TimBrooks, Jianfeng Wang, Linjie Li, † LongOuyang, † JuntangZhuang, † JoyceLee, † YufeiGuo, † WesamManassra, † PrafullaDhariwal, † CaseyChu, † YunxinJiao, and Aditya Ramesh.
\newblock Improving image generation with better captions.
\newblock \url{https://api.semanticscholar.org/CorpusID:264403242}.

\bibitem[Chen et~al.(2024)Chen, Ge, Xie, Wu, Yao, Ren, Wang, Luo, Lu, and Li]{pixartsigma}
Junsong Chen, Chongjian Ge, Enze Xie, Yue Wu, Lewei Yao, Xiaozhe Ren, Zhongdao Wang, Ping Luo, Huchuan Lu, and Zhenguo Li.
\newblock Pixart-$\sigma$: Weak-to-strong training of diffusion transformer for 4k text-to-image generation, 2024.

\bibitem[Chen et~al.(2025)Chen, Wu, Liu, Pan, Liu, Xie, Yu, and Ruan]{janus-pro}
Xiaokang Chen, Zhiyu Wu, Xingchao Liu, Zizheng Pan, Wen Liu, Zhenda Xie, Xingkai Yu, and Chong Ruan.
\newblock Janus-pro: Unified multimodal understanding and generation with data and model scaling, 2025.
\newblock \url{https://arxiv.org/abs/2501.17811}.

\bibitem[Creswell et~al.(2022)Creswell, Shanahan, and Higgins]{cot-2}
Antonia Creswell, Murray Shanahan, and Irina Higgins.
\newblock Selection-inference: Exploiting large language models for interpretable logical reasoning, 2022.
\newblock \url{https://arxiv.org/abs/2205.09712}.

\bibitem[DeepSeek-AI et~al.(2025)DeepSeek-AI, Guo, Yang, Zhang, Song, Zhang, Xu, Zhu, Ma, Wang, Bi, Zhang, Yu, Wu, Wu, Gou, Shao, Li, Gao, and et~al.]{deepseek-r1}
DeepSeek-AI, Daya Guo, Dejian Yang, Haowei Zhang, Junxiao Song, Ruoyu Zhang, Runxin Xu, Qihao Zhu, Shirong Ma, Peiyi Wang, Xiao Bi, Xiaokang Zhang, Xingkai Yu, Yu~Wu, Z.~F. Wu, Zhibin Gou, Zhihong Shao, Zhuoshu Li, Ziyi Gao, and Aixin~Liu et~al.
\newblock Deepseek-r1: Incentivizing reasoning capability in llms via reinforcement learning, 2025.
\newblock \url{https://arxiv.org/abs/2501.12948}.

\bibitem[Deng et~al.(2025)Deng, Zhu, Li, Gou, Li, Wang, Zhong, Yu, Nie, Song, Shi, and Fan]{bagel}
Chaorui Deng, Deyao Zhu, Kunchang Li, Chenhui Gou, Feng Li, Zeyu Wang, Shu Zhong, Weihao Yu, Xiaonan Nie, Ziang Song, Guang Shi, and Haoqi Fan.
\newblock Emerging properties in unified multimodal pretraining.
\newblock \emph{arXiv preprint arXiv:2505.14683}, 2025.

\bibitem[Dong et~al.(2024)Dong, Han, Peng, Qi, Ge, Yang, Zhao, Sun, Zhou, Wei, Kong, Zhang, Ma, and Yi]{dreamllm}
Runpei Dong, Chunrui Han, Yuang Peng, Zekun Qi, Zheng Ge, Jinrong Yang, Liang Zhao, Jianjian Sun, Hongyu Zhou, Haoran Wei, Xiangwen Kong, Xiangyu Zhang, Kaisheng Ma, and Li~Yi.
\newblock Dreamllm: Synergistic multimodal comprehension and creation, 2024.
\newblock \url{https://arxiv.org/abs/2309.11499}.

\bibitem[Esser et~al.(2024)Esser, Kulal, Blattmann, Entezari, Müller, Saini, Levi, Lorenz, Sauer, Boesel, Podell, Dockhorn, English, Lacey, Goodwin, Marek, and Rombach]{sd3medium}
Patrick Esser, Sumith Kulal, Andreas Blattmann, Rahim Entezari, Jonas Müller, Harry Saini, Yam Levi, Dominik Lorenz, Axel Sauer, Frederic Boesel, Dustin Podell, Tim Dockhorn, Zion English, Kyle Lacey, Alex Goodwin, Yannik Marek, and Robin Rombach.
\newblock Scaling rectified flow transformers for high-resolution image synthesis, 2024.
\newblock \url{https://arxiv.org/abs/2403.03206}.

\bibitem[Fang et~al.(2025)Fang, Duan, Wang, Huang, Li, Yan, Tian, Zeng, Zhao, Dai, Liu, and Li]{got}
Rongyao Fang, Chengqi Duan, Kun Wang, Linjiang Huang, Hao Li, Shilin Yan, Hao Tian, Xingyu Zeng, Rui Zhao, Jifeng Dai, Xihui Liu, and Hongsheng Li.
\newblock Got: Unleashing reasoning capability of multimodal large language model for visual generation and editing, 2025.
\newblock \url{https://arxiv.org/abs/2503.10639}.

\bibitem[Feng et~al.(2020)Feng, Chen, Lin, Wang, Yan, and Ren]{cot-3}
Yanlin Feng, Xinyue Chen, Bill~Yuchen Lin, Peifeng Wang, Jun Yan, and Xiang Ren.
\newblock Scalable multi-hop relational reasoning for knowledge-aware question answering.
\newblock In \emph{EMNLP}, pages 1295--1309, November 2020.

\bibitem[Ge et~al.(2025)Ge, Zhao, Zhu, Ge, Yi, Song, Li, Ding, and Shan]{seedx}
Yuying Ge, Sijie Zhao, Jinguo Zhu, Yixiao Ge, Kun Yi, Lin Song, Chen Li, Xiaohan Ding, and Ying Shan.
\newblock Seed-x: Multimodal models with unified multi-granularity comprehension and generation, 2025.
\newblock \url{https://arxiv.org/abs/2404.14396}.

\bibitem[Ghosh et~al.(2023)Ghosh, Hajishirzi, and Schmidt]{geneval}
Dhruba Ghosh, Hanna Hajishirzi, and Ludwig Schmidt.
\newblock Geneval: An object-focused framework for evaluating text-to-image alignment, 2023.
\newblock \url{https://arxiv.org/abs/2310.11513}.

\bibitem[Guo et~al.(2025)Guo, Zhang, Tong, Zhao, Huang, Zhang, Zhang, Liu, Zhang, Gao, Li, and Heng]{image-cot}
Ziyu Guo, Renrui Zhang, Chengzhuo Tong, Zhizheng Zhao, Rui Huang, Haoquan Zhang, Manyuan Zhang, Jiaming Liu, Shanghang Zhang, Peng Gao, Hongsheng Li, and Pheng-Ann Heng.
\newblock Can we generate images with cot? let's verify and reinforce image generation step by step, 2025.
\newblock \url{https://arxiv.org/abs/2501.13926}.

\bibitem[Hu et~al.(2024)Hu, Shi, Fu, Roth, Ostendorf, Zettlemoyer, Smith, and Krishna]{sketchvisual}
Yushi Hu, Weijia Shi, Xingyu Fu, Dan Roth, Mari Ostendorf, Luke Zettlemoyer, Noah~A Smith, and Ranjay Krishna.
\newblock Visual sketchpad: Sketching as a visual chain of thought for multimodal language models, 2024.
\newblock \url{https://arxiv.org/abs/2406.09403}.

\bibitem[Huang et~al.(2025)Huang, Jia, Zhai, Cao, Ye, Zhao, Xu, Hu, and Lin]{vision-r1}
Wenxuan Huang, Bohan Jia, Zijie Zhai, Shaosheng Cao, Zheyu Ye, Fei Zhao, Zhe Xu, Yao Hu, and Shaohui Lin.
\newblock Vision-r1: Incentivizing reasoning capability in multimodal large language models, 2025.
\newblock \url{https://arxiv.org/abs/2503.06749}.

\bibitem[Jiang et~al.(2025)Jiang, Guo, Zhang, Zong, Li, Zhuo, Yan, Heng, and Li]{t2i-r1}
Dongzhi Jiang, Ziyu Guo, Renrui Zhang, Zhuofan Zong, Hao Li, Le~Zhuo, Shilin Yan, Pheng-Ann Heng, and Hongsheng Li.
\newblock T2i-r1: Reinforcing image generation with collaborative semantic-level and token-level cot, 2025.
\newblock \url{https://arxiv.org/abs/2505.00703}.

\bibitem[Jung et~al.(2025)Jung, Kim, Kim, Lee, Kim, and Chang]{object-relation-cot}
Ji~Hyeok Jung, Eun~Tae Kim, Seoyeon Kim, Joo~Ho Lee, Bumsoo Kim, and Buru Chang.
\newblock Is 'right' right? enhancing object orientation understanding in multimodal large language models through egocentric instruction tuning, 2025.
\newblock \url{https://arxiv.org/abs/2411.16761}.

\bibitem[Labs et~al.(2025)Labs, Batifol, Blattmann, Boesel, Consul, Diagne, Dockhorn, English, English, Esser, Kulal, Lacey, Levi, Li, Lorenz, Müller, Podell, Rombach, Saini, Sauer, and Smith]{flux-kontext}
Black~Forest Labs, Stephen Batifol, Andreas Blattmann, Frederic Boesel, Saksham Consul, Cyril Diagne, Tim Dockhorn, Jack English, Zion English, Patrick Esser, Sumith Kulal, Kyle Lacey, Yam Levi, Cheng Li, Dominik Lorenz, Jonas Müller, Dustin Podell, Robin Rombach, Harry Saini, Axel Sauer, and Luke Smith.
\newblock Flux.1 kontext: Flow matching for in-context image generation and editing in latent space, 2025.
\newblock \url{https://arxiv.org/abs/2506.15742}.

\bibitem[Li et~al.(2025{\natexlab{a}})Li, Wu, Zhang, Xia, Mao, Dong, Vulić, and Wei]{mvot}
Chengzu Li, Wenshan Wu, Huanyu Zhang, Yan Xia, Shaoguang Mao, Li~Dong, Ivan Vulić, and Furu Wei.
\newblock Imagine while reasoning in space: Multimodal visualization-of-thought, 2025{\natexlab{a}}.
\newblock \url{https://arxiv.org/abs/2501.07542}.

\bibitem[Li et~al.(2024)Li, Kamko, Akhgari, Sabet, Xu, and Doshi]{playground-v2.5}
Daiqing Li, Aleks Kamko, Ehsan Akhgari, Ali Sabet, Linmiao Xu, and Suhail Doshi.
\newblock Playground v2.5: Three insights towards enhancing aesthetic quality in text-to-image generation, 2024.
\newblock \url{https://arxiv.org/abs/2402.17245}.

\bibitem[Li et~al.(2025{\natexlab{b}})Li, Bigverdi, Gu, Ma, Yang, Li, Choi, and Krishna]{spatial-cot}
Linjie Li, Mahtab Bigverdi, Jiawei Gu, Zixian Ma, Yinuo Yang, Ziang Li, Yejin Choi, and Ranjay Krishna.
\newblock Unfolding spatial cognition: Evaluating multimodal models on visual simulations, 2025{\natexlab{b}}.
\newblock \url{https://arxiv.org/abs/2506.04633}.

\bibitem[Liao et~al.(2025)Liao, Yang, Li, Li, Lin, Cheng, and Wang]{imagegen-cot}
Jiaqi Liao, Zhengyuan Yang, Linjie Li, Dianqi Li, Kevin Lin, Yu~Cheng, and Lijuan Wang.
\newblock Imagegen-cot: Enhancing text-to-image in-context learning with chain-of-thought reasoning, 2025.
\newblock \url{https://arxiv.org/abs/2503.19312}.

\bibitem[Liu et~al.(2025)Liu, Yan, Zaharia, and Abbeel]{lwm}
Hao Liu, Wilson Yan, Matei Zaharia, and Pieter Abbeel.
\newblock World model on million-length video and language with blockwise ringattention, 2025.
\newblock \url{https://arxiv.org/abs/2402.08268}.

\bibitem[Liu et~al.(2022)Liu, Gong, and Liu]{flow}
Xingchao Liu, Chengyue Gong, and Qiang Liu.
\newblock Flow straight and fast: Learning to generate and transfer data with rectified flow.
\newblock \emph{arXiv preprint arXiv:2209.03003}, 2022.

\bibitem[Ma et~al.(2025)Ma, Liu, Chen, Liu, Wu, Wu, Pan, Xie, Zhang, yu, Zhao, Wang, Liu, and Ruan]{janusflow}
Yiyang Ma, Xingchao Liu, Xiaokang Chen, Wen Liu, Chengyue Wu, Zhiyu Wu, Zizheng Pan, Zhenda Xie, Haowei Zhang, Xingkai yu, Liang Zhao, Yisong Wang, Jiaying Liu, and Chong Ruan.
\newblock Janusflow: Harmonizing autoregression and rectified flow for unified multimodal understanding and generation, 2025.
\newblock \url{https://arxiv.org/abs/2411.07975}.

\bibitem[Mitra et~al.(2024)Mitra, Huang, Darrell, and Herzig]{mcot}
Chancharik Mitra, Brandon Huang, Trevor Darrell, and Roei Herzig.
\newblock Compositional chain of thought prompting for large multimodal models.
\newblock In \emph{CVPR}, 2024.

\bibitem[Mu et~al.(2023)Mu, Zhang, Hu, Wang, Ding, Jin, Wang, Dai, Qiao, and Luo]{embodiedgpt}
Yao Mu, Qinglong Zhang, Mengkang Hu, Wenhai Wang, Mingyu Ding, Jun Jin, Bin Wang, Jifeng Dai, Yu~Qiao, and Ping Luo.
\newblock Embodiedgpt: Vision-language pre-training via embodied chain of thought, 2023.
\newblock \url{https://arxiv.org/abs/2305.15021}.

\bibitem[Niu et~al.(2025)Niu, Ning, Zheng, Jin, Lin, Jin, Liao, Feng, Ning, Zhu, and Yuan]{wise}
Yuwei Niu, Munan Ning, Mengren Zheng, Weiyang Jin, Bin Lin, Peng Jin, Jiaqi Liao, Chaoran Feng, Kunpeng Ning, Bin Zhu, and Li~Yuan.
\newblock Wise: A world knowledge-informed semantic evaluation for text-to-image generation, 2025.
\newblock \url{https://arxiv.org/abs/2503.07265}.

\bibitem[OpenAI et~al.(2024)OpenAI, :, Jaech, Kalai, Lerer, Richardson, El-Kishky, Low, Helyar, Madry, Beutel, Carney, Iftimie, Karpenko, Passos, Neitz, Prokofiev, and et~al.]{openaio1}
OpenAI, :, Aaron Jaech, Adam Kalai, Adam Lerer, Adam Richardson, Ahmed El-Kishky, Aiden Low, Alec Helyar, Aleksander Madry, Alex Beutel, Alex Carney, Alex Iftimie, Alex Karpenko, Alex~Tachard Passos, Alexander Neitz, Alexander Prokofiev, and Alexander~Wei et~al.
\newblock Openai o1 system card, 2024.
\newblock \url{https://arxiv.org/abs/2412.16720}.

\bibitem[Pan et~al.(2025)Pan, Shukla, Singh, Zhao, Mishra, Wang, Xu, Chen, Li, Juefei-Xu, Hou, and Xie]{metaquery}
Xichen Pan, Satya~Narayan Shukla, Aashu Singh, Zhuokai Zhao, Shlok~Kumar Mishra, Jialiang Wang, Zhiyang Xu, Jiuhai Chen, Kunpeng Li, Felix Juefei-Xu, Ji~Hou, and Saining Xie.
\newblock Transfer between modalities with metaqueries, 2025.
\newblock \url{https://arxiv.org/abs/2504.06256}.

\bibitem[Podell et~al.(2023)Podell, English, Lacey, Blattmann, Dockhorn, Müller, Penna, and Rombach]{sdxl}
Dustin Podell, Zion English, Kyle Lacey, Andreas Blattmann, Tim Dockhorn, Jonas Müller, Joe Penna, and Robin Rombach.
\newblock Sdxl: Improving latent diffusion models for high-resolution image synthesis, 2023.
\newblock \url{https://arxiv.org/abs/2307.01952}.

\bibitem[Qin et~al.(2025)Qin, Gong, Sun, Li, Yang, Yang, Qu, Tan, and Li]{unicot}
Luozheng Qin, Jia Gong, Yuqing Sun, Tianjiao Li, Mengping Yang, Xiaomeng Yang, Chao Qu, Zhiyu Tan, and Hao Li.
\newblock Uni-cot: Towards unified chain-of-thought reasoning across text and vision, 2025.
\newblock \url{https://arxiv.org/abs/2508.05606}.

\bibitem[Qu et~al.(2025)Qu, Zhang, Liu, Wang, Jiang, Gao, Ye, Du, Yuan, and Wu]{tokenflowxl}
Liao Qu, Huichao Zhang, Yiheng Liu, Xu~Wang, Yi~Jiang, Yiming Gao, Hu~Ye, Daniel~K. Du, Zehuan Yuan, and Xinglong Wu.
\newblock Tokenflow: Unified image tokenizer for multimodal understanding and generation, 2025.
\newblock \url{https://arxiv.org/abs/2412.03069}.

\bibitem[Ramesh et~al.(2022)Ramesh, Dhariwal, Nichol, Chu, and Chen]{dalle2}
Aditya Ramesh, Prafulla Dhariwal, Alex Nichol, Casey Chu, and Mark Chen.
\newblock Hierarchical text-conditional image generation with clip latents, 2022.
\newblock \url{https://arxiv.org/abs/2204.06125}.

\bibitem[Rombach et~al.(2022{\natexlab{a}})Rombach, Blattmann, Lorenz, Esser, and Ommer]{sd1.5}
Robin Rombach, Andreas Blattmann, Dominik Lorenz, Patrick Esser, and Björn Ommer.
\newblock High-resolution image synthesis with latent diffusion models, 2022{\natexlab{a}}.
\newblock \url{https://arxiv.org/abs/2112.10752}.

\bibitem[Rombach et~al.(2022{\natexlab{b}})Rombach, Blattmann, Lorenz, Esser, and Ommer]{sd21}
Robin Rombach, Andreas Blattmann, Dominik Lorenz, Patrick Esser, and Björn Ommer.
\newblock High-resolution image synthesis with latent diffusion models, 2022{\natexlab{b}}.
\newblock \url{https://arxiv.org/abs/2112.10752}.

\bibitem[Shi et~al.(2025)Shi, Han, Zhou, Liang, Lin, Zettlemoyer, and Yu]{llamafusion}
Weijia Shi, Xiaochuang Han, Chunting Zhou, Weixin Liang, Xi~Victoria Lin, Luke Zettlemoyer, and Lili Yu.
\newblock Lmfusion: Adapting pretrained language models for multimodal generation, 2025.
\newblock \url{https://arxiv.org/abs/2412.15188}.

\bibitem[Team(2025)]{chameleon}
Chameleon Team.
\newblock Chameleon: Mixed-modal early-fusion foundation models, 2025.
\newblock \url{https://arxiv.org/abs/2405.09818}.

\bibitem[Tong et~al.(2024)Tong, Fan, Zhu, Xiong, Chen, Sinha, Rabbat, LeCun, Xie, and Liu]{metamorph}
Shengbang Tong, David Fan, Jiachen Zhu, Yunyang Xiong, Xinlei Chen, Koustuv Sinha, Michael Rabbat, Yann LeCun, Saining Xie, and Zhuang Liu.
\newblock Metamorph: Multimodal understanding and generation via instruction tuning, 2024.
\newblock \url{https://arxiv.org/abs/2412.14164}.

\bibitem[van~den Oord et~al.(2018)van~den Oord, Vinyals, and Kavukcuoglu]{vq-vae}
Aaron van~den Oord, Oriol Vinyals, and Koray Kavukcuoglu.
\newblock Neural discrete representation learning, 2018.
\newblock \url{https://arxiv.org/abs/1711.00937}.

\bibitem[Wang et~al.(2024{\natexlab{a}})Wang, Lu, Yang, Huang, Han, Hou, Zhang, and Xu]{Illume}
Chunwei Wang, Guansong Lu, Junwei Yang, Runhui Huang, Jianhua Han, Lu~Hou, Wei Zhang, and Hang Xu.
\newblock Illume: Illuminating your llms to see, draw, and self-enhance, 2024{\natexlab{a}}.
\newblock \url{https://arxiv.org/abs/2412.06673}.

\bibitem[Wang et~al.(2024{\natexlab{b}})Wang, Zhang, Luo, Sun, Cui, Wang, Zhang, Wang, Li, Yu, Zhao, Ao, Min, Li, Wu, Zhao, Zhang, Wang, Liu, He, Yang, Liu, Lin, Huang, and Wang]{emu3}
Xinlong Wang, Xiaosong Zhang, Zhengxiong Luo, Quan Sun, Yufeng Cui, Jinsheng Wang, Fan Zhang, Yueze Wang, Zhen Li, Qiying Yu, Yingli Zhao, Yulong Ao, Xuebin Min, Tao Li, Boya Wu, Bo~Zhao, Bowen Zhang, Liangdong Wang, Guang Liu, Zheqi He, Xi~Yang, Jingjing Liu, Yonghua Lin, Tiejun Huang, and Zhongyuan Wang.
\newblock Emu3: Next-token prediction is all you need, 2024{\natexlab{b}}.
\newblock \url{https://arxiv.org/abs/2409.18869}.

\bibitem[Wang and Zhou(2024)]{cot-prompt}
Xuezhi Wang and Denny Zhou.
\newblock Chain-of-thought reasoning without prompting, 2024.
\newblock \url{https://arxiv.org/abs/2402.10200}.

\bibitem[Wei et~al.(2023)Wei, Wang, Schuurmans, Bosma, Ichter, Xia, Chi, Le, and Zhou]{cot}
Jason Wei, Xuezhi Wang, Dale Schuurmans, Maarten Bosma, Brian Ichter, Fei Xia, Ed~Chi, Quoc Le, and Denny Zhou.
\newblock Chain-of-thought prompting elicits reasoning in large language models, 2023.
\newblock \url{https://arxiv.org/abs/2201.11903}.

\bibitem[Wu et~al.(2024{\natexlab{a}})Wu, Chen, Wu, Ma, Liu, Pan, Liu, Xie, Yu, Ruan, and Luo]{janus}
Chengyue Wu, Xiaokang Chen, Zhiyu Wu, Yiyang Ma, Xingchao Liu, Zizheng Pan, Wen Liu, Zhenda Xie, Xingkai Yu, Chong Ruan, and Ping Luo.
\newblock Janus: Decoupling visual encoding for unified multimodal understanding and generation, 2024{\natexlab{a}}.
\newblock \url{https://arxiv.org/abs/2410.13848}.

\bibitem[Wu et~al.(2024{\natexlab{b}})Wu, Fei, Qu, Ji, and Chua]{nextgpt}
Shengqiong Wu, Hao Fei, Leigang Qu, Wei Ji, and Tat-Seng Chua.
\newblock Next-gpt: Any-to-any multimodal llm, 2024{\natexlab{b}}.
\newblock \url{https://arxiv.org/abs/2309.05519}.

\bibitem[Wu et~al.(2025{\natexlab{a}})Wu, Zhang, Chen, Tang, Li, Fang, Zhu, Xie, Yin, Yi, Han, and Lu]{vila-u}
Yecheng Wu, Zhuoyang Zhang, Junyu Chen, Haotian Tang, Dacheng Li, Yunhao Fang, Ligeng Zhu, Enze Xie, Hongxu Yin, Li~Yi, Song Han, and Yao Lu.
\newblock Vila-u: a unified foundation model integrating visual understanding and generation, 2025{\natexlab{a}}.
\newblock \url{https://arxiv.org/abs/2409.04429}.

\bibitem[Wu et~al.(2025{\natexlab{b}})Wu, Wang, Ye, Du, Jegelka, and Wang]{cot-less}
Yuyang Wu, Yifei Wang, Ziyu Ye, Tianqi Du, Stefanie Jegelka, and Yisen Wang.
\newblock When more is less: Understanding chain-of-thought length in llms, 2025{\natexlab{b}}.
\newblock \url{https://arxiv.org/abs/2502.07266}.

\bibitem[Xie et~al.(2025{\natexlab{a}})Xie, Mao, Bai, Zhang, Wang, Lin, Gu, Chen, Yang, and Shou]{showo}
Jinheng Xie, Weijia Mao, Zechen Bai, David~Junhao Zhang, Weihao Wang, Kevin~Qinghong Lin, Yuchao Gu, Zhijie Chen, Zhenheng Yang, and Mike~Zheng Shou.
\newblock Show-o: One single transformer to unify multimodal understanding and generation, 2025{\natexlab{a}}.
\newblock \url{https://arxiv.org/abs/2408.12528}.

\bibitem[Xie et~al.(2025{\natexlab{b}})Xie, Yang, and Shou]{showo2}
Jinheng Xie, Zhenheng Yang, and Mike~Zheng Shou.
\newblock Show-o2: Improved native unified multimodal models, 2025{\natexlab{b}}.
\newblock \url{https://arxiv.org/abs/2506.15564}.

\bibitem[Zhang et~al.(2025)Zhang, Li, Yang, Wang, Yang, Qi, Bao, Chen, Luo, and Qiu]{reasongen-r1}
Yu~Zhang, Yunqi Li, Yifan Yang, Rui Wang, Yuqing Yang, Dai Qi, Jianmin Bao, Dongdong Chen, Chong Luo, and Lili Qiu.
\newblock Reasongen-r1: Cot for autoregressive image generation models through sft and rl, 2025.
\newblock \url{https://arxiv.org/abs/2505.24875}.

\bibitem[Zheng et~al.(2023)Zheng, Yang, Tang, Zhou, and Yang]{mcot-2}
Ge~Zheng, Bin Yang, Jiajin Tang, Hong-Yu Zhou, and Sibei Yang.
\newblock Ddcot: Duty-distinct chain-of-thought prompting for multimodal reasoning in language models.
\newblock In \emph{NeurIPS2023}, 2023.

\bibitem[Zhou et~al.(2024{\natexlab{a}})Zhou, Yu, Babu, Tirumala, Yasunaga, Shamis, Kahn, Ma, Zettlemoyer, and Levy]{transfusion}
Chunting Zhou, Lili Yu, Arun Babu, Kushal Tirumala, Michihiro Yasunaga, Leonid Shamis, Jacob Kahn, Xuezhe Ma, Luke Zettlemoyer, and Omer Levy.
\newblock Transfusion: Predict the next token and diffuse images with one multi-modal model, 2024{\natexlab{a}}.
\newblock \url{https://arxiv.org/abs/2408.11039}.

\bibitem[Zhou et~al.(2024{\natexlab{b}})Zhou, Zhou, Hu, Lu, Gao, and Zhang]{image-of-thought}
Qiji Zhou, Ruochen Zhou, Zike Hu, Panzhong Lu, Siyang Gao, and Yue Zhang.
\newblock Image-of-thought prompting for visual reasoning refinement in multimodal large language models, 2024{\natexlab{b}}.
\newblock \url{https://arxiv.org/abs/2405.13872}.

\bibitem[Zhuo et~al.(2025)Zhuo, Zhao, Paul, Liao, Zhang, Xin, Gao, Elhoseiny, and Li]{from-reflection-to-perfection}
Le~Zhuo, Liangbing Zhao, Sayak Paul, Yue Liao, Renrui Zhang, Yi~Xin, Peng Gao, Mohamed Elhoseiny, and Hongsheng Li.
\newblock From reflection to perfection: Scaling inference-time optimization for text-to-image diffusion models via reflection tuning.
\newblock In \emph{ICCV}, 2025.

\end{thebibliography}



\end{document}